\documentclass{article}
\usepackage{spconf,amsmath,epsfig}
\usepackage{xcolor}
\usepackage{graphicx}
\usepackage{multirow}
\usepackage{bbm}
\usepackage{amssymb}
\usepackage{amsfonts}
\usepackage{amsmath}
\usepackage{hyperref}

\title{One for all: Toward unified Foundation Models for Earth vision}
\name{Zhitong Xiong, Yi Wang, Fahong Zhang, Xiao Xiang Zhu}
\address{ Chair of Data Science in Earth Observation, Technical University of Munich, Munich, Germany}

\begin{document}
%
\maketitle
\begin{abstract}
Foundation models characterized by extensive parameters and trained on large-scale datasets have demonstrated remarkable efficacy across various downstream tasks for remote sensing data. Current remote sensing foundation models typically specialize in a single modality or a specific spatial resolution range, limiting their versatility for downstream datasets. While there have been attempts to develop multi-modal remote sensing foundation models, they typically employ separate vision encoders for each modality or spatial resolution, necessitating a switch in backbones contingent upon the input data. To address this issue, we introduce a simple yet effective method, termed OFA-Net (One-For-All Network): employing a single, shared Transformer backbone for multiple data modalities with different spatial resolutions. Using the masked image modeling mechanism, we pre-train a single Transformer backbone on a curated multi-modal dataset with this simple design. Then the backbone model can be used in different downstream tasks, thus forging a path towards a unified foundation backbone model in Earth vision. The proposed method is evaluated on 12 distinct downstream tasks and demonstrates promising performance.
\end{abstract}
\section{Introduction}
\label{sec:intro}
Multiple satellites in orbit provide invaluable data essential for understanding and managing our planet. The richness of the Earth observation data lies in its diversity, encompassing various modalities such as optical, radar, multispectral, hyperspectral, and thermal imagery, each offering unique insights \cite{zhu2017deep} \cite{xiong2022earthnets}. This multiplicity is crucial in applications ranging from environmental monitoring~\cite{zhao2024causal} to urban planning~\cite{xiong2023benchmark}, demonstrating the indispensable role of remote sensing in Earth sciences.

The advent of foundation models has currently revolutionized the processing and analysis of remote sensing data \cite{zhu2024foundations,guo2023skysense, bastani2023satlaspretrain,cha2023billion,cong2022satmae,wang2022ssl4eo,xiong2024neural}. Characterized by their extensive parameters and pre-trained on large-scale datasets, these models have greatly enhanced the performance on different downstream tasks. Despite their successes, current foundation models in remote sensing exhibit a critical limitation: they are predominantly tailored to either a single data modality \cite{cong2022satmae} or a specific range of spatial resolutions \cite{smith2023earthpt} \cite{jain2022self}. This specialization constrains their applicability across the diverse spectrum of remote sensing datasets, limiting their potential and flexibility in broader, more complex applications.

\begin{figure}[h]
	\centering
	\includegraphics[width=0.475\textwidth]{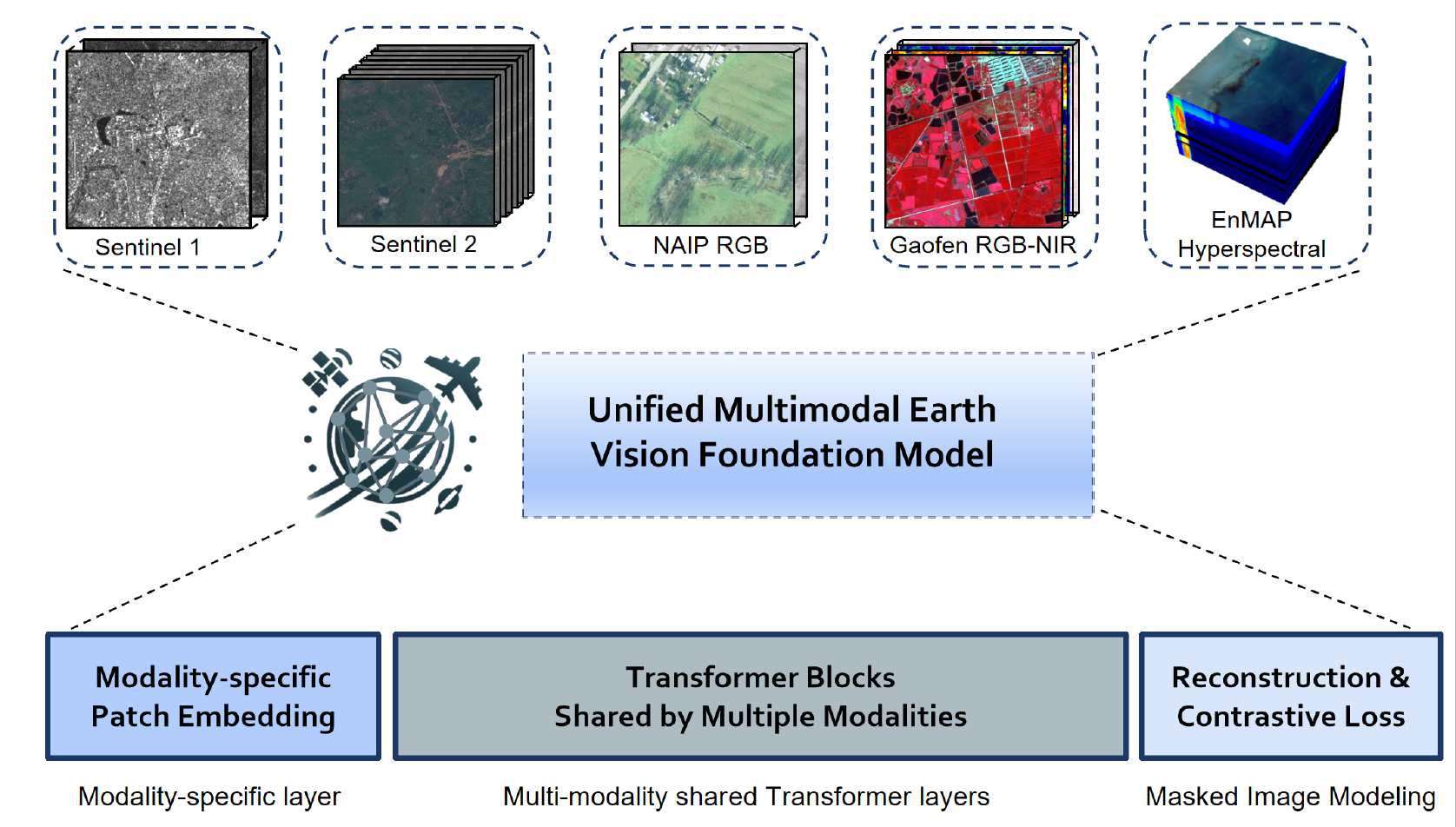}
	\caption{Illustration of the proposed method. Our model is designed to handle input data from a range of modalities, and varying spatial resolutions, such as 30 meters and 1 meter, using a singular, unified framework. This integrative approach allows for the simultaneous processing of all modalities within one comprehensive model. }
	\label{fig1}
\end{figure}	

In response, the remote sensing community has tried to develop multi-modal foundation models. However, these models typically rely on separate vision encoders for either each modality or spatial resolution. Such an approach necessitates the switching of backbones based on different input data, hindering flexibility and operational efficiency in downstream applications.
\begin{figure*}[h]
	\centering
	\includegraphics[width=0.92\textwidth]{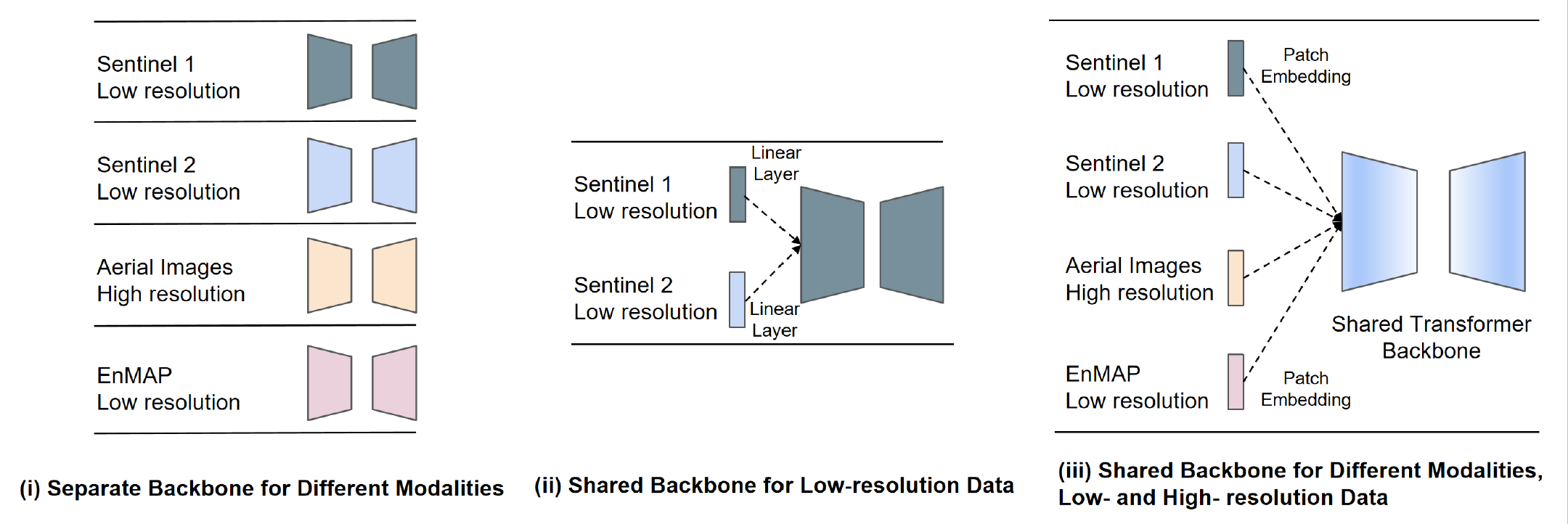}
	\caption{Illustration of existing and the proposed foundation models for multi-modal data. }
	\label{fig2}
\end{figure*}	
In this context, a model capable of seamlessly integrating multiple data modalities and spatial resolutions within a single framework could dramatically enhance the adaptability and efficiency of remote sensing data understanding.

In this work, we propose a simple approach to this challenge: a unified foundation model employing a single, shared vision Transformer \cite{dosovitskiy2020image} backbone. This model accommodates data with various modalities and spatial resolutions, aiming to eliminate the need for multiple specialized models, as illustrated in Fig. \ref{fig1}. 

As shown in Fig. \ref{fig2} (iii), the proposed OFA-Net is designed to handle input data from a range of modalities and varying spatial resolutions, such as 30 meters and 1 meter, using a singular, unified vision Transformer. This integrative approach allows for the simultaneous processing of all modalities within one comprehensive model. which is different from conventional methods. We pre-train this Transformer backbone on a meticulously curated multi-modal dataset, leveraging the masked image modeling mechanism to enhance its adaptability. Our approach offers a simpler yet effective solution for Earth vision tasks. We validate our model on 12 downstream tasks in the GEO-Bench dataset \cite{lacoste2023geo}, demonstrating its robustness and versatility. 

\begin{figure*}[h]
    \centering
    \includegraphics[width=0.93\textwidth]{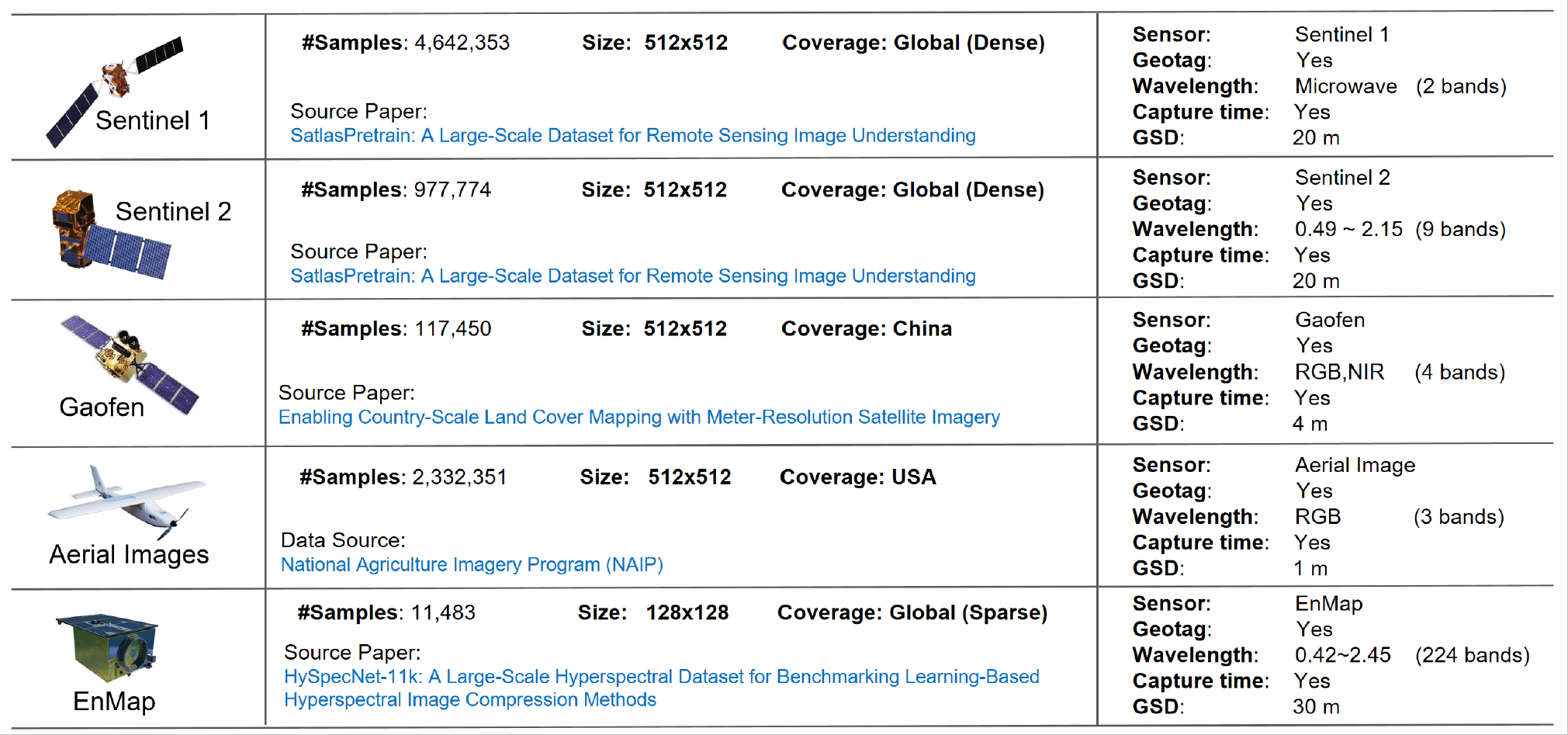}
    \caption{Detailed information of the five sub-datasets in the curated multi-modal dataset.}
    \label{fig_data}
\end{figure*}	
\section{Methodology}
\label{sec:method}
\subsection{Multi-modal Dataset Construction}
As shown in Fig. \ref{fig_data}, we have constructed an extensive multi-modal dataset designed to underpin the development of unified foundation models for Earth vision. This dataset is composed of five distinct modalities, each offering unique spectral and spatial data characteristics:

\textbf{Sentinel-1}: The Sentinel-1 dataset includes 4,642,353 samples of Synthetic Aperture Radar (SAR) imagery, with a spatial resolution of about $5\times 20$ meters. Each image captures two bands (vv and vh) and is 512x512 pixels in size, providing dense global coverage.

\textbf{Sentinel-2}: Comprising 977,774 multispectral imagery samples, this dataset has a spectral range with nine bands, from 0.49 to 2.15 $\mu m$, maintaining a spatial resolution of 10 meters with each image sized at 512x512 pixels for dense global coverage. We use the Sentinel-2 data collected and processed by \cite{bastani2023satlaspretrain}.

\textbf{Gaofen}: To include images from the Gaofen satellite, We use the dataset collected by \cite{tong2023enabling}. We crop 117,450 image patches of 512$\times$512 pixel resolution from the dataset. Each image includes four bands encompassing RGB and NIR wavelengths with a spatial resolution of around 4 meters. This dataset mainly covers different cities in China. 

\textbf{NAIP}: For high-resolution optical images, we use the dataset collected and processed by \cite{bastani2023satlaspretrain}. This dataset includes 2,332,351 high-resolution aerial images from the National Agriculture Imagery Program (NAIP), covering the USA with a fine spatial resolution of approximately 1 meter and consisting of RGB images across three bands with a size of 512x512 pixels.

\textbf{EnMAP}: The multi-modal dataset is further enriched with 11,483 hyperspectral image samples from EnMAP, which is published in \cite{fuchs2023hyspecnet}. The hyperspectral images have a spatial resolution of 30 meters and capture a wide spectral range with 224 bands, each sized at 128x128 pixels.

\begin{figure*}[h]
    \centering
    \includegraphics[width=0.92\textwidth]{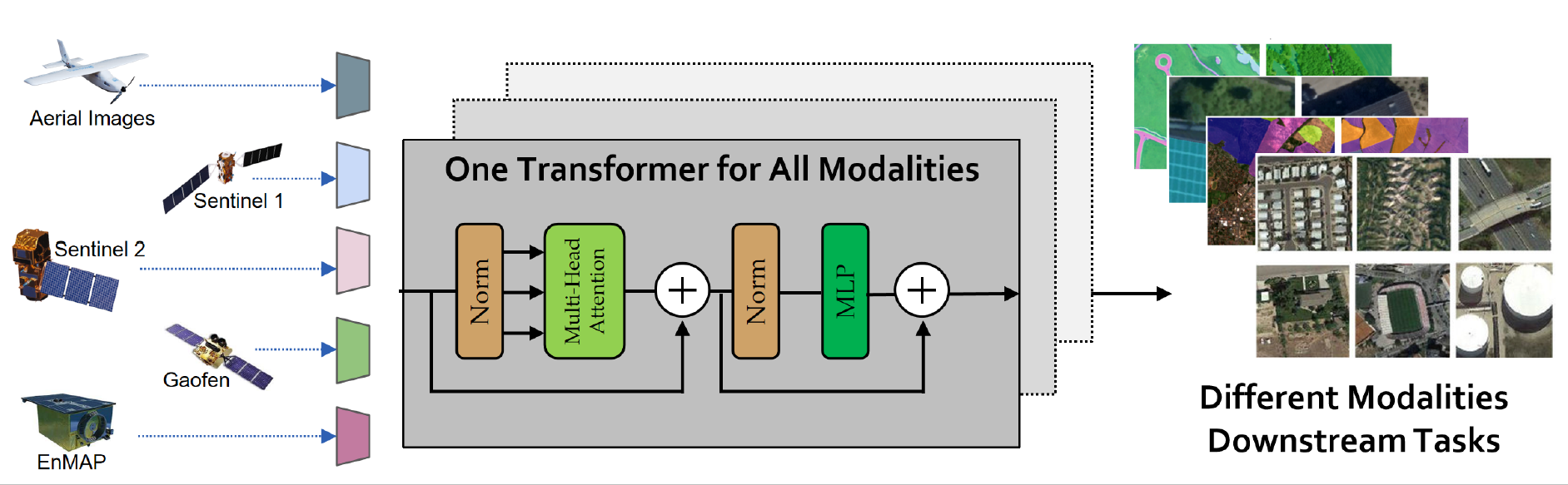}
    \caption{Workflow of the proposed unified foundation model for multiple data modalities.}
    \label{fig_ours}
\end{figure*}

As illustrated in Fig. \ref{fig_ours}, the proposed OFA-Net contains mainly two components: 1) the individual patch embedding layers, and 2) the shared Transformer backbone. This simple design enables the model to extract and process features from diverse remote sensing modalities without the need to train multiple foundation models.
\begin{table*}[ht]
\centering
\caption{Performance comparison of different methods on various classification datasets}
\label{tab:performance_comparison1}
\scalebox{0.95}{
\begin{tabular}{c|c|c|c|c|c|c}
\hline
\textbf{Methods} & \textbf{m-bigeathnet} & \textbf{m-forestnet} & \textbf{m-brick-kiln} & \textbf{m-pv4ger} & \textbf{m-so2sat} & \textbf{m-eurosat} \\ \hline
Random Init.              & 52.89             & 41.52                  & 84.51                & 91.32                 & 38.31             & 69.53              \\ \hline
MAE Single       & 55.41             & 42.95                  & 88.89                & 92.19                 & 44.42             & 78.00              \\ \hline
SatMAE \cite{cong2022satmae}    & 55.12           &---       & \textbf{91.89}     &---           & 45.59                 & 73.15 \\ \hline
\textbf{OFA-Net (ours)}         & 57.13             & \textbf{45.12}                  & {91.29}                & \textbf{93.19}  & \textbf{46.04}      & \textbf{81.00}    \\ \hline
\end{tabular}}
\end{table*}

\begin{table*}[ht]
\centering
\caption{Performance comparison of different methods on various segmentation datasets}
\label{tab:performance_comparison2}
\scalebox{0.84}{
\begin{tabular}{c|c|c|c|c|c|c}
\hline
\textbf{Methods} & \textbf{m-pv4ger-seg} & \textbf{m-nz-cattle} & \textbf{m-NeonTree} & \textbf{m-cashew-plantation} & \textbf{m-SA-crop-type} & \textbf{m-chesapeake-landcover} \\ \hline
Random Init.              & 81.63             & 74.12                 & 51.27                & 27.65               & 29.11             & 47.16              \\ \hline
MAE Single       & 88.43             & 76.40                 & \textbf{52.99}                & 29.42               & 30.67             & 51.90           \\ \hline
\textbf{OFA-Net (Ours)}          & \textbf{89.43}             & \textbf{77.63}  & {52.64}          & \textbf{37.39}    & \textbf{31.98}    & \textbf{54.50}            \\  \hline
\end{tabular}}
\end{table*}
\textbf{Individual Patch Embedding Layer}.
The first component of the model comprises separate patch embedding layers tailored to each modality. There exist inherent differences in input channels across modalities. SAR images from Sentinel-1 are with two bands. Hyperspectral images from EnMAP are with 224 bands. Hence, it is imperative to have a specialized embedding process that can effectively translate the raw pixel data into a format suitable for the Transformer backbone. 

Let $P_{s1},P_{s2},P_{naip},P_{g}$, and $P_{hyper}$ denote the patch embedding operations for five different modalities. After resizing the input images from different modalities $X_{s1}\in \mathbb{R}^{h,w,2},X_{s2}\in \mathbb{R}^{h,w,9}, X_{naip}\in \mathbb{R}^{h,w,3}, X_{g}\in \mathbb{R}^{h,w,4}, X_{h}\in \mathbb{R}^{h,w,224}$ into 224x224 pixels, we can compute the embeddings simply by $E=P(X)$ for each modality.
\label{sec:experiments}

\textbf{Shared Transformer Backbone.} The second component is a shared Transformer backbone. In this work, this single Transformer architecture processes the embedded patches from all modalities with different spatial resolutions. This shared backbone can learn a generalized representation that is flexible and robust enough to handle not only the variety of data modalities but also their diverse spatial resolutions. In this process, we input the embeddings for each modality $E_{s1}\in \mathbb{R}^{n,d},E_{s2}\in \mathbb{R}^{n,d}, E_{naip}\in \mathbb{R}^{n,d}, E_{g}\in \mathbb{R}^{n,d}, E_{h}\in \mathbb{R}^{h,w,224}$ to the Transformer backbone $F_b$ to learn deep features. Here $n$ represents the number of tokens and $d$ denotes the feature dimension.

\textbf{Masked Image Modeling.}
For the model training, we utilize the masked image modeling-based self-supervised learning loss. Specifically, we employ different decoders for different data modalities to reconstruct the randomly masked parts of the inputs. A significant advantage introduced by masked image modeling is its inherent design that does not necessitate spatially aligned multi-modal datasets. Traditional multi-modal learning approaches often rely on the precise alignment of different modalities, which can be a challenging and expensive process given the variability in sensor acquisition parameters and conditions.

\section{Experiments}
\subsection{Downstream Datasets}
\label{section::datasets}
To assess the effectiveness of our proposed unified foundation model, we use the GEO-Bench benchmark datasets \cite{lacoste2023geo}, which encompass a diverse array of tasks pertinent to Earth vision. The GEO-Bench dataset includes 12 tasks, split evenly between image classification and segmentation, each representing a common challenge in remote sensing. 

\subsection{Experimental Settings}
\label{section::settings}
Considering the high computational cost, we use 10,000 data samples for each sub-dataset, with 50,000 samples in total to pre-train the OFA-Net for 100 epochs. We adopt linear probing as our primary evaluation strategy, known for its capacity to measure the quality of representations learned by self-supervised learning. This approach involves freezing the weights of the pre-trained model and training a linear classifier on top of the representations for each task. By doing so, we can directly evaluate the discriminative power of the learned features without fully fine-tuning the entire model, thus providing insight into the model's ability to generalize. A learning rate of 1e-2 is used for all the datasets. For the classification tasks, we use top-1 Accuracy as the evaluation metric. For segmentation tasks, we use mIoU as the metric. A learning rate of 1e-4 is used for all the datasets. All the experiments are conducted using PyTorch on four NVIDIA A6000 GPUs each with 48GB Memory.

\subsection{Comparison Experiments}
To objectively evaluate our model, we provide the performance of four different methods on six downstream classification datasets under the linear probing setting. The results are presented in Table \ref{tab:performance_comparison1}. Analyzing the table, OFA-Net generally outperforms the other methods across most datasets, indicating the benefit of multimodal pretraining in enhancing the model's feature extraction capabilities and generalization. The improvement is particularly notable on \textbf{m-forestnet} and \textbf{m-so2sat} datasets, where the OFA-Net method exceeds the performance of a randomly initialized model by approximately 3.6\% and 7.7\%, respectively. The MAE Single method shows consistent improvement over random initialization, which highlights the advantage of pretraining on a single data modality compared to no pretraining. However, the gains from MAE Single are less than those from OFA-Net, underscoring the added value of multimodal learning.

Table \ref{tab:performance_comparison2} presents performance for segmentation tasks on various datasets, comparing methods that utilize a Vision Transformer (ViT) with different pretraining strategies: Random Initialization (Random Init.), pretraining on single modalities (MAE Single), and pretraining on multiple modalities (OFA-Net). On the \textbf{m-NeonTree} dataset, OFA-Net shows a modest improvement over MAE Single. On the other five datasets, OFA-Net demonstrates superior performance when compared to the other methods. This suggests that pretraining on multiple modalities provides a more comprehensive feature representation, leading to more accurate segmentation. MAE Single outperforms Random Initialization in all cases, which aligns with the expected outcome that pretraining can significantly enhance the model's ability to generalize and accurately segment images. 

\section{Conclusion}
In this work, we introduce a simple yet effective method, the OFA-Net, for unified foundation models for remote sensing data. The OFA-Net consists of a single, shared Transformer backbone and dedicated patch embedding layers for multiple data modalities with different spatial resolutions. The model is trained using the masked image modeling mechanism on a carefully curated multi-modal dataset with five distinct modalities. Then the backbone model is evaluated in different downstream tasks. The experimental results on 12 different downstream tasks show that our simple method demonstrates promising performance over foundation models trained using single modalities.

\section{Acknowledgement}
This work is jointly supported by the German Federal Ministry of Education and Research (BMBF) in the framework of the international future AI lab ``AI4EO -- Artificial Intelligence for Earth Observation: Reasoning, Uncertainties, Ethics and Beyond'' (grant number: 01DD20001), by German Federal Ministry for Economic Affairs and Climate Action in the framework of the ``national center of excellence ML4Earth'' (grant number: 50EE2201C), and by the German Federal Ministry for the Environment, Nature Conservation, Nuclear Safety and Consumer Protection (BMUV) based on a resolution of the German Bundestag (grant number: 67KI32002B; Acronym: \textit{EKAPEx}).

\bibliographystyle{IEEEbib}
\bibliography{strings,refs}

\end{document}